\documentclass[pdflatex,sn-mathphys-num]{sn-jnl}

\usepackage{graphicx}%
\usepackage{multirow}%
\usepackage{amsmath,amssymb,amsfonts}%
\usepackage{amsthm}%
\usepackage{mathrsfs}%
\usepackage[title]{appendix}%
\usepackage{xcolor}%
\usepackage{textcomp}%
\usepackage{manyfoot}%
\usepackage{booktabs}%
\usepackage{algorithm}%
\usepackage{algorithmicx}%
\usepackage{algpseudocode}%
\usepackage{listings}%
\usepackage{makecell}
\usepackage{newtxtext}
\usepackage{newtxmath}
\usepackage{booktabs}
\usepackage{caption}
\usepackage{caption}
\usepackage{varwidth}
\usepackage{listings}
\usepackage{xcolor}
\usepackage{makecell}
\usepackage{array}
\usepackage{tabularx}
\usepackage{graphicx}
\usepackage{pdflscape}
\usepackage{longtable}
\usepackage{appendix}
\usepackage{siunitx}
\usepackage{float}

\theoremstyle{thmstyleone}%
\theoremstyle{thmstyletwo}%

\theoremstyle{thmstylethree}%

\begin{document}

\title[Article Title]{Can News Predict the Market? Limits of Zero-Shot Financial NLP and the Role of Explainable AI}


\author{\fnm{Ali} \sur{M Karaoglu}}

\author*{\fnm{Shreyank} \sur{N Gowda}}\email{shreyank.narayanagowda@nottingham.ac.uk}

\affil*{\orgdiv{School of Computer Science}, \orgname{University of Nottingham}, \orgaddress{\city{Nottingham}, \country{United Kingdom}}}


\abstract{Can financial news reliably predict short-term stock movements? Despite advances in large language models, this question remains unresolved. We revisit this problem using a zero-shot natural language processing framework, investigating whether models can extract actionable signals from financial news without domain-specific training. We design a structured pipeline that combines zero-shot natural language inference with temporal aggregation, explicitly modelling recency and event-dependent impact horizons when integrating information across articles. To address the need for transparency in high-stakes settings, we introduce a multi-layered explainability framework that links predictions to token-level, article-level, and aggregate evidence, and produces grounded natural language rationales. Across multiple models and prediction horizons, we find that zero-shot approaches consistently fail to outperform simple baselines, with particularly weak performance on negative movements, suggesting deeper structural limitations in mapping news sentiment to short-term price dynamics. However, explainability signals reliably distinguish between trustworthy and unreliable predictions, offering practical value even when accuracy is limited. These findings highlight the limits of zero-shot financial NLP and motivate a shift toward decision-support systems that prioritise transparency and uncertainty awareness. Code: \url{https://github.com/alimert05/zero-shot-stock-xai}}




\maketitle

\section{Introduction}

Financial markets react rapidly to new information, and financial news remains one of the most widely used textual sources for interpreting short-term market sentiment \citep{Fama1970, SchumakerChen2009}. However, converting news into reliable stock movement predictions is difficult. Articles are noisy, frequently duplicated across outlets, unevenly relevant to the target company, and often written in cautious or implicit language where sentiment is not stated directly \citep{Evans2020NewsProvenance, VanDeKauterBreeschHoste2015, JacobsHoste2021ImplicitSentiment}. These issues are especially important in short-horizon forecasting, where even small amounts of irrelevant or stale information can distort the aggregate signal.

A further challenge is temporal alignment. News publication and market response do not always occur on the same timescale. Articles may be released outside market hours or over weekends, and different event types can produce immediate, delayed, or gradual reactions \citep{DakalbabKumarAbuTalibNasir2025}. For example, earnings announcements may trigger rapid price responses, while regulatory developments, restructuring, or financial distress may unfold over longer periods. Standard pipelines that aggregate all articles within a fixed lookback window risk mixing short-lived and slow-moving signals, reducing their relevance to the chosen prediction horizon.

Most existing financial NLP pipelines also depend on supervised or domain-adapted models trained on manually annotated or market-labelled datasets \citep{MaloEtAl2014PhraseBank, Ashtiani2023, DuEtAl2024}. Although these models can perform well on benchmark sentiment tasks, they require labelled data, may need periodic retraining, and can struggle to generalise across changing market conditions \citep{HinderVaquetHammer2024, BayramAhmed2025}. Zero-shot language models offer an attractive alternative because they can classify financial text without task-specific fine-tuning \citep{FatourosSoldatosKouroumaliMakridisKyriazis2023}. However, applying zero-shot models directly to stock prediction remains risky, since prompt sensitivity, input noise, and weak temporal structure can turn plausible sentiment outputs into unreliable market signals.

This paper investigates whether zero-shot financial NLP can support short-horizon stock movement prediction when embedded within a structured, temporally aware, and explainable pipeline. Rather than treating news sentiment as a direct trading signal, we model each prediction as an evidence aggregation problem. Articles are filtered for company relevance, weighted by recency and estimated event-dependent impact horizon, and then aggregated into a three-way prediction over positive, negative, and neutral movement. Since financial prediction is a high-stakes setting where accuracy alone is insufficient, the pipeline also provides multi-level explanations that connect each prediction to token-level cues, article-level influence, aggregate evidence quality, and natural language rationales \citep{YeoVanDerHeeverMaoCambriaSatapathyMengaldo2025}.

The central research question is:

\begin{quote}
\textit{Can an impact-aware, zero-shot approach deliver accurate and trustworthy predictions of short-term stock market movements from financial news, without requiring domain-specific fine-tuning?}
\end{quote}

We study this question across three dimensions. First, we evaluate whether zero-shot NLI-based prediction with temporal weighting can match or exceed domain-adapted and general-purpose alternatives at the pipeline level. Second, we assess whether a multi-layered explanation framework can provide grounded and stable rationales for each prediction. Third, we examine whether evidence-quality diagnostics can identify trustworthy and unreliable predictions, even when directional accuracy is limited.

The contributions of this paper are as follows:

\begin{enumerate}
    \item We develop an impact-aware zero-shot financial NLP pipeline for short-horizon stock movement prediction, combining company relevance filtering, recency weighting, event-dependent impact-horizon weighting, and three-way abstention-based prediction.

    \item We introduce a multi-layered explainability framework that links predictions to token-level attributions, article-level contributions, contrastive evidence, robustness checks, evidence-quality diagnostics, and grounded natural language summaries.

    \item We evaluate the approach on a 480-case dataset covering 20 US-listed companies across six prediction horizons, comparing the proposed zero-shot NLI configuration against domain-adapted and general-purpose alternatives.

    \item We show that zero-shot financial NLP remains limited for short-horizon directional prediction, particularly for negative movements, but that structured explainability and evidence-quality diagnostics can provide useful decision-support value by distinguishing more reliable predictions from weaker ones.
\end{enumerate}

\section{Related Work}
\label{sec:related-work}

\subsection{Financial News and Market Prediction}

Financial news has long been recognised as an important source of market information. While the Efficient Market Hypothesis suggests that publicly available information should be rapidly incorporated into prices \citep{Fama1970}, numerous studies have shown that market reactions are often delayed, incomplete, or dependent on the type of information released \citep{ArmstrongCardellaSabah2021, Fink2021}. Earnings announcements, product recalls, regulatory actions, and macroeconomic events have all been linked to significant price movements, although the magnitude and duration of these effects vary considerably \citep{UnsalHassanZirek2017, LiuShankarYun2017, KawasDockery2023}.

A recurring challenge is that financial news is not a homogeneous signal. News articles differ in relevance, novelty, publication timing, and expected market impact. Furthermore, media outlets frequently republish similar stories, causing article volume to overestimate genuinely new information \citep{Evans2020NewsProvenance}. Prior work also shows that market reactions occur on different timescales depending on the event type and the way information is presented \citep{BriereEtAl2023, ChenFenglerHaerdleLiu2022}. These findings suggest that effective prediction systems should account for both article relevance and temporal dynamics rather than relying solely on aggregated sentiment scores.

\subsection{Financial Language Models}

Recent advances in financial NLP have been driven by domain-adapted language models. FinBERT adapts BERT to the financial domain and demonstrates strong performance on financial sentiment benchmarks such as Financial PhraseBank and FiQA \citep{AraciFinBERT2019}. More recently, FinGPT extends this idea using parameter-efficient fine-tuning of generative language models on large collections of financial news \citep{YangLiuWangFinGPT2023}. Representative results are shown in Tables~\ref{tab:finbert-results} and \ref{tab:fingpt-results}.

\begin{table}[!htbp]
    \caption{FinBERT results on FPB (classification) and FiQA Task 1 (regression). Adapted from \citet{AraciFinBERT2019}.}
    \label{tab:finbert-results}
    \centering
    \begin{tabular*}{\textwidth}{@{\extracolsep{\fill}}lcc@{}}
        \toprule
        \textit{Dataset / setting} & \textit{Metric(s)} & \textit{FinBERT result} \\
        \midrule
        FPB (all data) & Accuracy, Macro-F1 & 0.86, 0.84 \\
        FPB (100\% agreement) & Accuracy, Macro-F1 & 0.97, 0.95 \\
        FiQA Task 1 & MSE, $R^2$ & 0.07, 0.55 \\
        \bottomrule
    \end{tabular*}
\end{table}

\begin{table}[!htbp]
    \caption{Sentiment classification performance (3-way). Adapted from \citet{YangLiuWangFinGPT2023}.}
    \label{tab:fingpt-results}
    \centering
    \begin{tabular*}{\textwidth}{@{\extracolsep{\fill}}lccccc@{}}
        \toprule
        \textit{Model} & \textit{Acc.} & \textit{Macro-F1} & \textit{Pos-F1} & \textit{Neg-F1} & \textit{Neu-F1} \\
        \midrule
        ChatGPT (0-shot) & 63.4 & 61.7 & 64.0 & 59.1 & 62.0 \\
        Llama3.1-8B (0-shot) & 57.9 & 54.4 & 56.1 & 53.2 & 54.0 \\
        FinBERT & 71.2 & 69.9 & 73.0 & 69.1 & 67.5 \\
        FinGPT (LoRA-SFT) & 78.8 & 77.3 & 79.6 & 76.8 & 75.4 \\
        FinGPT (SFT+RLSP) & \textbf{82.1} & \textbf{80.9} & \textbf{83.4} & \textbf{81.5} & \textbf{77.8} \\
        \bottomrule
    \end{tabular*}
\end{table}

These approaches demonstrate that financial adaptation can substantially improve sentiment classification performance. However, both depend on labelled or market-derived training data and may require retraining as market conditions evolve \citep{HinderVaquetHammer2024, BayramAhmed2025}. This motivates interest in zero-shot alternatives.

\subsection{Zero-Shot Financial NLP}

Zero-shot methods aim to perform classification without task-specific training \citep{AlhoshanFerrariZhao2023}. In finance, this is attractive because labelled datasets are costly to construct and may not generalise across market regimes \citep{ZuoJiangZhou2024, BashiriJirkovskyHernandezLeon2025}. More broadly, zero-shot learning has demonstrated strong transfer capabilities across vision and multimodal tasks, although performance often depends heavily on the quality of semantic representations and the alignment between training and deployment domains \citep{zerodiff, zsl1, zsl2, zsl3, spot}.

Existing work has explored both prompt-based generative LLMs and NLI-based classification. Prompt-based approaches can achieve competitive sentiment classification performance under carefully designed prompts, but results are often sensitive to prompt wording and evaluation protocol \citep{FatourosSoldatosKouroumaliMakridisKyriazis2023, LiChanZhuPeiMaLiuShah2023, ShenZhang2024FSA}. Table~\ref{tab:zeroshot-summary} summarises representative findings.

\begin{table}[!htbp]
\caption{Representative zero-shot sentiment results reported in the literature.}
\label{tab:zeroshot-summary}
\centering
\begin{tabular}{lccc}
\toprule
Study & Model & Setting & F1 \\
\midrule
Fatouros et al. & GPT & Financial sentiment & 0.79 \\
Li et al. & GPT-4 & Financial PhraseBank & 0.96 \\
Shen \& Zhang & GPT-4o & Financial PhraseBank & 0.84 \\
Yang et al. & ChatGPT & Market-labelled sentiment & 0.62 \\
\bottomrule
\end{tabular}
\end{table}

An alternative is NLI-based classification, where labels are expressed as hypotheses and the model predicts the degree of entailment between the input text and each candidate hypothesis \citep{YinHayRoth2019}. Models such as DeBERTa and RoBERTa achieve strong performance on NLI benchmarks and provide deterministic, computationally efficient inference without requiring task-specific training \citep{HeGaoChen2021DeBERTaV3, LiuOttGoyalDuJoshiChenLevyLewisZettlemoyerStoyanov2019}. However, these models are not inherently adapted to financial language and may require supporting mechanisms such as relevance filtering and structured aggregation to handle noisy financial text \citep{LaurerVanAtteveldtCasasWelbers2023, JacobsHoste2021ImplicitSentiment}.

\subsection{Explainability in Finance}

The increasing use of AI in finance has led to growing interest in explainability, particularly in applications where model outputs influence investment decisions and risk management \citep{YeoVanDerHeeverMaoCambriaSatapathyMengaldo2025}. Traditional techniques such as LIME and SHAP provide local feature attributions, but these explanations can be difficult to interpret when predictions are derived from collections of articles rather than single inputs \citep{RibeiroSinghGuestrin2016LIME, LundbergLee2017SHAP}.

Recent work has therefore explored broader explanation strategies, including article-level evidence selection, robustness analysis, and natural-language explanation generation using LLMs \citep{Jang2025SelectiveNewsSelection, KoaMaNgChua2024}. While these methods improve interpretability, concerns remain regarding faithfulness and hallucination, particularly when explanations are generated independently of the model's actual reasoning process \citep{TurpinMichaelPerez2023, JiLeeFrieske2023}. Existing evidence suggests that effective financial explanations should be grounded in the underlying evidence and provide transparency at multiple levels, including token-level, article-level, and prediction-level analysis.

\subsection{Research Gap}

The literature highlights three unresolved challenges. First, financial news exhibits complex temporal behaviour, yet many prediction systems treat all articles within a fixed window equally. Second, while zero-shot methods remove the need for labelled training data, they are rarely evaluated as complete prediction pipelines operating over collections of articles and multiple prediction horizons. Third, existing explainability approaches typically focus on either local attribution or narrative explanation, rather than combining multiple forms of evidence into a unified trust assessment.

To address these limitations, this work investigates an impact-aware zero-shot prediction pipeline that combines temporal weighting, NLI-based classification, and multi-layered explainability. Rather than treating sentiment classification as the end goal, we examine whether zero-shot financial NLP can support transparent and trustworthy decision-making in short-horizon stock prediction.


\section{Methodology}
\label{sec:methodology}

\begin{figure}[!htbp]
    \centering
    \includegraphics[width=1\linewidth]{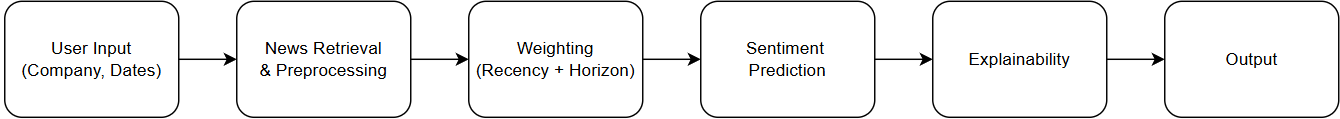}
    \caption{Overview of the proposed zero-shot financial news prediction and explainability pipeline.}
    \label{fig:full_pipe}
\end{figure}

We propose an impact-aware zero-shot prediction pipeline that takes a company name and prediction window as input, retrieves relevant financial news, assigns temporally informed article weights, predicts stock movement direction, and produces an explainability report with a reliability rating. The pipeline has four stages: data acquisition, preprocessing and filtering, prediction, and explainability, as shown in Figure~\ref{fig:full_pipe}.

\subsection{Problem Formulation}

The task is formulated as three-way classification over positive, negative, and neutral stock movement. We use a three-way formulation rather than binary up/down prediction because many short-horizon price changes fall within normal market noise. An explicit neutral class allows the system to abstain from directional predictions when the evidence is weak or ambiguous.

For a company $c$ and prediction window $W$, the system retrieves articles published before the prediction start date, filters them for relevance, assigns each retained article a sentiment distribution and temporal weight, and aggregates the weighted evidence into a final prediction. The output is not intended as an automated trading signal, but as a decision-support summary that exposes both the prediction and the evidence behind it.

\subsection{Data Acquisition and Temporal Alignment}

The user-provided company name is first resolved to a stock ticker using the Yahoo Finance search API. News articles are then retrieved from Finnhub, which provides structured metadata and article summaries. The lookback period is computed dynamically as:

\begin{equation}
    \textit{backward\_days} =
    \min\left(90, \max\left(7, \left\lceil 5\sqrt{W} \right\rceil\right)\right),
\end{equation}

where $W$ is the prediction window in days. This gives short windows enough recent coverage while preventing longer windows from retrieving excessive stale news. The lookback period extends only backwards from the prediction start date, ensuring that no article published during or after the prediction period is used as input.

Each article is assigned to a market trading session. Article timestamps are converted from UTC to US Eastern Time, and a 4 PM ET market-close cutoff is applied. Articles published before market close are assigned to that trading day, while articles published after close, on weekends, or on holidays are rolled forward to the next valid trading session using the \texttt{exchange\_calendars} library. This ensures that recency and impact-horizon calculations reflect when the market could reasonably react to the news.

\subsection{Preprocessing and Relevance Filtering}

The retrieved articles are filtered in three stages. First, title-based deduplication removes repeated versions of the same story across outlets, retaining the oldest version or the longer version if timestamps tie. The number of outlets carrying the same headline is preserved as a coverage count for later weighting.

Second, article-level entity matching removes articles that do not mention the target company or ticker in the headline or body. This is necessary because ticker-based news APIs can return articles that mention unrelated entities or contain only weak company relevance.

Third, sentence-level relevance filtering removes boilerplate, subscription prompts, legal disclaimers, and unrelated content from otherwise relevant articles. This is implemented using DeBERTa in a zero-shot NLI formulation. Each sentence is paired with a company-specific relevance hypothesis:

\begin{quote}
\textit{This sentence is relevant/irrelevant to \{Company\} (\{Ticker\}), including products/brands, financial performance, guidance, operations, lawsuits, regulation, risks, and major announcements.}
\end{quote}

Sentences with relevance scores below 0.5 are discarded. If no sentence passes the threshold, the highest-scoring sentence is retained as a safeguard against reducing useful articles to headline-only inputs.

\subsection{Impact-Aware Article Weighting}
\label{sec:sec4}

Each retained article receives a final weight that combines recency, event-dependent impact horizon, source coverage, and headline-only discounting.

\subsubsection{Recency Weighting}

The recency weight is defined as:

\begin{equation}
    w_{\text{recency}} = \lambda^d,
\end{equation}

\begin{figure}[!htbp]
    \centering
    \includegraphics[width=0.9\linewidth]{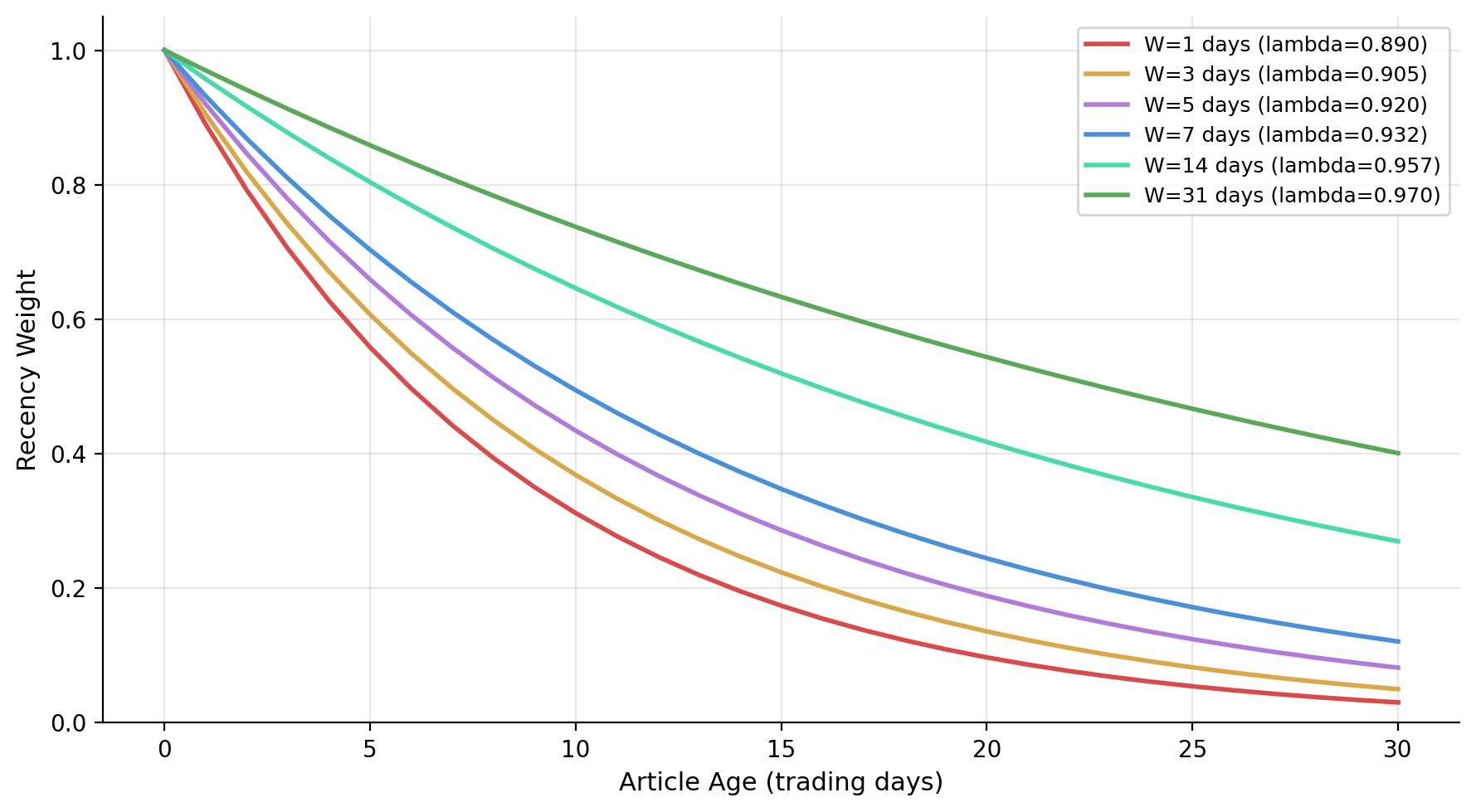}
    \caption{Recency decay curves for different prediction horizons. Short-horizon forecasts apply more aggressive temporal discounting, while longer-horizon forecasts retain older articles for longer.}
    \label{fig:recency_decay}
\end{figure}

where $d$ is the number of trading days between the article's market date and the prediction start date. The decay parameter $\lambda$ is interpolated from horizon-specific anchor values tuned on the validation set. Figure~\ref{fig:recency_decay} illustrates the resulting decay profiles. Short prediction horizons discount older articles aggressively, while longer horizons preserve a broader evidence base. This reflects the intuition that information useful for next-day forecasting becomes stale much more quickly than information relevant to longer prediction windows.

\subsubsection{Impact Horizon Classification}

Different financial events affect markets on different timescales. To model this, each article is classified into one of nine event families using the zero-shot NLI hypothesis:

\begin{quote}
\textit{The primary firm-specific event in this news article is \{\}.}
\end{quote}

Each event family is assigned a primary and secondary impact horizon based on the event-study literature. These horizons are summarised in Table~\ref{tab:event_horizons}.

\begin{table}[!htbp]
\caption{Event families and their primary and secondary market reaction horizons.}
\label{tab:event_horizons}
\centering
\renewcommand{\arraystretch}{1.15}
\setlength{\tabcolsep}{6pt}
\begin{tabularx}{\textwidth}{
>{\raggedright\arraybackslash\bfseries}p{3.4cm}
>{\raggedright\arraybackslash}p{2.7cm}
>{\raggedright\arraybackslash}p{2.7cm}
>{\raggedright\arraybackslash}X}
\toprule
\textit{Event Family} & \textit{Primary Horizon} & \textit{Secondary Horizon} & \textit{Key Reference} \\
\midrule
Earnings, guidance, financial results & 1 day & 6--10 days & \citep{BallBrown1968, BernardThomas1989} \\
Analyst upgrade/downgrade & 1 day & 2--5 days & \citep{KimLinSlovin1997, Womack1996} \\
Product launch, partnership, contract & 2--5 days & 6--10 days & \citep{WarrenSorescu2017} \\
Buyback, dividend, stock offering & 2--5 days & 11--20 days & \citep{Vermaelen1981} \\
Lawsuit, regulatory action & 2--5 days & 11--20 days & \citep{KarpoffLott1993} \\
M\&A, restructuring & 6--10 days & 21--31 days & \citep{Schwert1996} \\
CEO/executive change & 2--5 days & 6--10 days & \citep{DenisDenis1995} \\
Market commentary, opinion & 1 day & 2--5 days & \citep{HestonSinha2016} \\
Financial distress, credit downgrade & 21--31 days & 11--20 days & \citep{HolthausenLeftwich1986, CampbellHilscherSzilagyi2008} \\
\bottomrule
\end{tabularx}
\end{table}

\subsubsection{Impact Horizon Weighting}

The horizon weight measures how well an article's expected impact timing aligns with the prediction window:

\begin{equation}
    w_{\text{horizon}} =
    \exp\left(
    -\frac{(\textit{impact\_day} - \mu)^2}{2\sigma^2}
    \right),
\end{equation}

where $\textit{impact\_day} = \textit{horizon\_days} - \textit{days\_ago}$, $\mu = W/2$, and $\sigma = \max(W/2, 3.0)$. The Gaussian form gives highest weight to articles whose expected impact falls near the centre of the prediction window, while still allowing partially aligned articles to contribute. A minimum value of 0.05 is applied so that articles are not completely removed due to possible event-classification uncertainty.

Primary and secondary horizon weights are blended using classifier confidence:

\begin{equation}
    w_{\text{combined\_horizon}} =
    m \cdot w_{\text{primary}} + (1-m) \cdot w_{\text{secondary}},
\end{equation}

where $m = 0.60 + 0.25 \cdot \textit{confidence}$. This gives more influence to the primary horizon when event classification confidence is high, while retaining secondary-horizon information when confidence is lower.

\subsubsection{Final Article Weight}

The final article weight is:

\begin{equation}
\label{final_weight}
    w_i =
    \sqrt{w_{\text{recency}} \cdot w_{\text{combined\_horizon}}}
    \cdot \log_2(1 + \textit{coverage\_count})
    \cdot d_h,
\end{equation}

where $d_h = 0.5$ for headline-only articles and $d_h = 1$ otherwise. The geometric mean penalises articles that are strong on only one temporal dimension, while the logarithmic coverage factor gives moderate additional weight to widely syndicated stories without allowing them to dominate the aggregate.

\subsection{Zero-Shot Sentiment Prediction and Abstention}

The pipeline uses DeBERTa for relevance filtering and impact-horizon classification, and RoBERTa for final sentiment classification. The final sentiment hypothesis is:

\begin{quote}
\textit{The implication of this statement is \{\}.}
\end{quote}

with candidate labels \textit{bullish}, \textit{bearish}, and \textit{neutral}. Each article input is prefixed with the target company name to provide explicit entity context.

The NLI model is used in multi-label sigmoid mode rather than softmax mode. Sigmoid scoring treats each label independently, avoiding artificial competition between labels when the article is genuinely ambiguous. Article-level label scores are aggregated using the final weights in Equation~\ref{final_weight}. A directional prediction is issued only when the corresponding score exceeds a tuned threshold:

\begin{equation}
    \text{label} =
    \begin{cases}
    \text{positive}, & s_{\text{pos}} \geq \tau_{\text{pos}} \text{ and } (s_{\text{pos}} \geq s_{\text{neg}} \text{ or } s_{\text{neg}} < \tau_{\text{neg}}), \\
    \text{negative}, & s_{\text{neg}} \geq \tau_{\text{neg}} \text{ and } (s_{\text{neg}} \geq s_{\text{pos}} \text{ or } s_{\text{pos}} < \tau_{\text{pos}}), \\
    \text{neutral}, & \text{otherwise}.
    \end{cases}
\end{equation}

Thresholds are tuned on the validation set to maximise macro-F1. The selected values are $\tau_{\text{pos}} = 0.56$ and $\tau_{\text{neg}} = 0.26$. Table~\ref{tab:threshold_tuning_roberta} shows that threshold tuning substantially improves macro-F1 by recovering neutral predictions that are lost under simple argmax classification.

\begin{table}[!htbp]
\centering
\caption{Comparison of performance before and after threshold tuning on the validation set.}
\label{tab:threshold_tuning_roberta}
\renewcommand{\arraystretch}{1}
\setlength{\tabcolsep}{8pt}
\begin{tabularx}{\textwidth}{
>{\raggedright\arraybackslash\bfseries}X
>{\raggedleft\arraybackslash}p{1.8cm}
>{\raggedleft\arraybackslash}p{1.8cm}
>{\raggedleft\arraybackslash}p{1.5cm}
>{\raggedleft\arraybackslash}p{1.5cm}
>{\raggedleft\arraybackslash}p{1.5cm}}
\toprule
\textit{Configuration} & \textit{Accuracy} & \textit{Macro-F1} & \textit{Neg-F1} & \textit{Neu-F1} & \textit{Pos-F1} \\
\midrule
Argmax, no thresholds & 0.2986 & 0.1870 & 0.0964 & 0.0000 & 0.4646 \\
Tuned thresholds & 0.3958 & 0.3353 & 0.1931 & 0.5627 & 0.2500 \\
\bottomrule
\end{tabularx}
\end{table}

\subsection{Explainability and Reliability Assessment}

The explainability framework is designed to answer three questions: what evidence supported the prediction, how robust was the prediction, and how much trust should be placed in it. It combines token-level, article-level, and pipeline-level explanations.

Token-level explanations are generated using LIME \citep{RibeiroSinghGuestrin2016LIME}, selected because it is model-agnostic and can explain the black-box NLI classifier without requiring access to internal gradients. For each prediction, LIME is applied to a small set of influential supporting and opposing articles.

Article-level explanation is provided through contrastive contribution analysis. For each article, the net contribution toward the winning label over the runner-up is:

\begin{equation}
    \text{net direction}_i =
    \frac{w_i^{\text{winner}} - w_i^{\text{runner-up}}}{\sum_j w_j}.
\end{equation}

This identifies which articles most strongly separate the chosen label from its closest alternative. Robustness is assessed using a greedy flip-set analysis: articles supporting the winning label are removed in descending order of influence until the prediction changes, or until no flip is possible. A small flip set indicates fragility, while a larger flip set indicates that the prediction is supported by broader evidence.

To make the explanation more readable, articles are also grouped into thematic storylines using TF-IDF vectors and agglomerative clustering. Cluster labels are generated using a locally hosted Llama model, with a keyword-based fallback when generation is unavailable. The same local model is used to produce a grounded natural-language summary from structured evidence, including influential articles, quality flags, and attribution outputs. The prompt is constrained to these inputs to reduce the risk of unsupported explanation generation.

Finally, the pipeline computes evidence-quality flags that identify common failure modes. These flags are summarised in Table~\ref{tab:quality_flags}.

\begin{table}[!htbp]
\caption{Quality and reliability flags used in the prediction pipeline.}
\label{tab:quality_flags}
\centering
\renewcommand{\arraystretch}{1}
\setlength{\tabcolsep}{8pt}
\begin{tabular*}{\textwidth}{@{\extracolsep{\fill}}>{\raggedright\arraybackslash}p{0.22\textwidth}>{\raggedright\arraybackslash}p{0.50\textwidth}>{\raggedright\arraybackslash}p{0.18\textwidth}@{}}
\toprule
\textit{Flag} & \textit{What it checks} & \textit{Threshold} \\
\midrule
\textbf{Thin evidence} & Too few articles to form a reliable aggregate. & $< 5$ articles \\
\addlinespace[0.4em]
\textbf{Weight concentration} & Prediction dominated by a small number of articles, measured using the Herfindahl--Hirschman Index. & $\mathrm{HHI} > 0.4$ \\
\addlinespace[0.4em]
\textbf{Label-margin} & Small gap between the top two sentiment scores. & $< 0.15$ \\
\addlinespace[0.4em]
\textbf{Source diversity} & Most articles come from the same editorial source. & $> 60\%$ from one source \\
\addlinespace[0.4em]
\textbf{Flip sensitivity} & Prediction can be changed by removing a small number of articles. & $\leq 5$ articles \\
\addlinespace[0.4em]
\textbf{Horizon coverage} & Retrieved articles do not adequately span the intended lookback window. & Coverage gap detected \\
\bottomrule
\end{tabular*}
\end{table}

Weight concentration is measured using the Herfindahl--Hirschman Index:

\begin{equation}
    \mathrm{HHI} =
    \sum_{i=1}^{N}
    \left(
    \frac{w_i}{\sum_j w_j}
    \right)^2.
\end{equation}

The overall reliability rating is assigned from the number of triggered flags: HIGH for 0 or 1 flags, MEDIUM for 2 flags, and LOW for 3 or more flags. This rating does not replace predictive accuracy, but provides a transparent indication of whether the evidence behind a prediction is broad, stable, and sufficiently diverse.


\section{Evaluation}
\label{sec:eva}

\subsection{Evaluation Setup}

We evaluate the proposed pipeline across five sentiment models, six prediction horizons, five sectors, and multiple explainability diagnostics. The compared sentiment models are RoBERTa, FinBERT, FinGPT, Llama 3.1 8B, and Mistral 7B. RoBERTa is the proposed NLI-based sentiment component, while the remaining models provide domain-adapted and general-purpose baselines. The evaluation addresses three questions: whether zero-shot news-based prediction can outperform simple baselines, how the proposed pipeline compares with alternative sentiment models, and whether the explainability layer can identify more trustworthy predictions.

The dataset contains 480 test cases generated from 20 US-listed companies across five sectors: Technology, Finance, Healthcare, Energy, and Consumer. Each company is evaluated across six prediction windows of 1, 3, 5, 7, 14, and 31 days, using four start dates between April 2025 and January 2026. This gives $20 \times 6 \times 4 = 480$ cases and avoids evaluating the system on a single market period.

Ground-truth labels are derived from stock price movements obtained using \texttt{yfinance}, independently of the news data. For each case, the percentage change between the closing price at the start and end of the prediction window determines the movement direction. To avoid applying a fixed threshold across stocks with different volatility levels, the neutral band is scaled using a point-in-time EWMA volatility estimate:

\begin{equation}
\label{ewma}
    \tau = k \cdot \sigma_{\mathrm{EWMA}} \cdot \sqrt{W},
\end{equation}

where $W$ is the prediction window in trading days and $k=0.5$. Returns within $[-\tau, \tau]$ are labelled neutral, returns above the band are labelled positive, and returns below the band are labelled negative. The volatility estimate is computed using only price data available before the prediction window, preventing future leakage.

The 480 cases are split into a 288-case tuning set and a 192-case holdout set at the ticker level, so that all cases for a given company appear in only one split. This reduces leakage from overlapping company-specific news coverage. The split closely preserves the overall class distribution, as shown in Table~\ref{tab:split_class_distribution}. Performance is reported using accuracy, macro-precision, macro-recall, macro-F1, and per-class F1. Macro-F1 is emphasised because the negative class is smaller but practically important.

\begin{table}[!htbp]
    \caption{Class distribution across the tuning and holdout splits.}
    \label{tab:split_class_distribution}
    \centering
    \renewcommand{\arraystretch}{1}
    \setlength{\tabcolsep}{8pt}
    \begin{tabular*}{\textwidth}{@{\extracolsep{\fill}}lcccccc@{}}
        \toprule
        \textit{Split} & \textit{Cases} & \textit{Articles} & \textit{Companies} & \textit{Positive} & \textit{Negative} & \textit{Neutral} \\
        \midrule
        \textbf{Tune} & 288 & 27,568 & 12 & 92 (31.9\%) & 56 (19.4\%) & 140 (48.6\%) \\
        \addlinespace[0.4em]
        \textbf{Holdout} & 192 & 25,200 & 8 & 61 (31.8\%) & 38 (19.8\%) & 93 (48.4\%) \\
        \addlinespace[0.4em]
        \textbf{Total} & 480 & 52,768 & 20 & 153 (31.9\%) & 94 (19.6\%) & 233 (48.5\%) \\
        \bottomrule
    \end{tabular*}
\end{table}

\subsection{Financial PhraseBank Sanity Check}
\label{sec:FPB_bench}

Before evaluating full pipeline performance, we test all five sentiment models on Financial PhraseBank \citep{MaloEtAl2014PhraseBank}. This acts as a sentence-level sanity check: if a model cannot identify sentiment in clean financial sentences, poor performance on the noisier multi-article prediction task would be unsurprising. Results across the 50\%, 75\%, and all-agree subsets are reported in Table~\ref{tab:fpb_results_roberta}.

\begin{table}[!htbp]
    \caption{Financial PhraseBank benchmark results across agreement splits.}
    \label{tab:fpb_results_roberta}
    \centering
    \renewcommand{\arraystretch}{1}
    \setlength{\tabcolsep}{8pt}
    \begin{tabular*}{\textwidth}{@{\extracolsep{\fill}}lccccc@{}}
        \toprule
        \textit{Model} & \textit{Macro-F1} & \textit{Accuracy} & \textit{Pos-F1} & \textit{Neg-F1} & \textit{Neu-F1} \\
        \midrule
        \multicolumn{6}{l}{\textbf{50agree} ($N = 4{,}846$)} \\
        RoBERTa & 0.6405 & 0.6112 & 0.5645 & 0.7572 & 0.5998 \\
        FinBERT & 0.8825 & 0.8896 & 0.8620 & 0.8786 & 0.9071 \\
        FinGPT & \textbf{0.9157} & \textbf{0.9199} & \textbf{0.8771} & \textbf{0.9344} & \textbf{0.9355} \\
        Llama 3.1 8B & 0.7868 & 0.8040 & 0.6687 & 0.8448 & 0.8470 \\
        Mistral 7B & 0.7837 & 0.8044 & 0.6812 & 0.8208 & 0.8491 \\
        \addlinespace[0.5em]
        \multicolumn{6}{l}{\textbf{75agree} ($N = 3{,}453$)} \\
        RoBERTa & 0.6635 & 0.6392 & 0.5618 & 0.7784 & 0.6503 \\
        FinBERT & 0.9365 & 0.9473 & 0.9305 & 0.9178 & 0.9611 \\
        FinGPT & \textbf{0.9648} & \textbf{0.9676} & \textbf{0.9495} & \textbf{0.9706} & \textbf{0.9743} \\
        Llama 3.1 8B & 0.8647 & 0.8787 & 0.7785 & 0.9078 & 0.9080 \\
        Mistral 7B & 0.8623 & 0.8795 & 0.7975 & 0.8800 & 0.9095 \\
        \addlinespace[0.5em]
        \multicolumn{6}{l}{\textbf{allagree} ($N = 2{,}264$)} \\
        RoBERTa & 0.6968 & 0.6727 & 0.5760 & 0.8265 & 0.6879 \\
        FinBERT & 0.9625 & 0.9717 & 0.9620 & 0.9430 & 0.9825 \\
        FinGPT & \textbf{0.9860} & \textbf{0.9872} & \textbf{0.9815} & \textbf{0.9869} & \textbf{0.9896} \\
        Llama 3.1 8B & 0.9147 & 0.9205 & 0.8508 & 0.9554 & 0.9381 \\
        Mistral 7B & 0.9095 & 0.9196 & 0.8655 & 0.9244 & 0.9387 \\
        \bottomrule
    \end{tabular*}
\end{table}

FinGPT and FinBERT achieve the strongest sentence-level results, confirming the advantage of domain adaptation on clean financial sentiment benchmarks. RoBERTa performs worse overall, but it achieves relatively strong negative-class F1 on the all-agree subset. This distinction is central to the later analysis: the model can recognise bearish language in isolated financial sentences, but full pipeline prediction requires mapping noisy, aggregated news evidence to future stock movement.

\subsection{Model Comparison on the Holdout Set}
\label{sec:model_comparison}

Table~\ref{tab:model_performance_issues} reports holdout performance for all five sentiment models using the full prediction pipeline. RoBERTa uses the tuned thresholding rule described in Section~\ref{sec:methodology}; the other models use argmax classification.

\begin{table}[!htbp]
    \caption{Model performance and issue counts on the holdout set.}
    \label{tab:model_performance_issues}
    \centering
    \renewcommand{\arraystretch}{1}
    \setlength{\tabcolsep}{8pt}
    \begin{tabular*}{\textwidth}{@{\extracolsep{\fill}}lcccccc@{}}
        \toprule
        \textit{Model} & \textit{Accuracy} & \textit{Macro-F1} & \textit{Neg-F1} & \textit{Neu-F1} & \textit{Pos-F1} & \textit{Issues} \\
        \midrule
        \textbf{RoBERTa} & 0.3750 & \textbf{0.3409} & \textbf{0.1667} & 0.5193 & 0.3368 & 0 \\
        \textbf{FinBERT} & 0.3594 & 0.2877 & 0.0364 & 0.4138 & 0.4130 & 0 \\
        \textbf{FinGPT} & \textbf{0.3802} & 0.2708 & 0.0400 & \textbf{0.5225} & 0.2500 & 16 \\
        \textbf{Llama 3.1 8B} & 0.3021 & 0.2472 & 0.0323 & 0.3611 & 0.3483 & 15 \\
        \textbf{Mistral 7B} & 0.3594 & 0.2830 & 0.0385 & 0.3649 & \textbf{0.4457} & 14 \\
        \bottomrule
    \end{tabular*}
\end{table}

No model exceeds the majority-class baseline of 48.4\%, indicating that short-horizon stock movement prediction from news sentiment alone remains difficult. RoBERTa achieves the highest macro-F1 despite weaker sentence-level performance on Financial PhraseBank, suggesting that benchmark sentiment accuracy does not directly translate to pipeline-level stock prediction. FinGPT achieves the highest accuracy, but its lower macro-F1 reflects weaker balance across the three classes.

The most consistent failure is negative-class detection. The four argmax-based models obtain negative F1 scores between 0.0323 and 0.0400, while RoBERTa improves negative F1 to 0.1667 through asymmetric decision thresholds. This suggests that negative movements are not merely underrepresented, but structurally difficult to identify from the available news evidence. The LLM-based models also produce 14 to 16 invalid or unparsable outputs, whereas the NLI-based models produce no such failures.

\subsection{Explainability and Reliability Evaluation}

The explainability layer is evaluated by asking whether it produces meaningful diagnostic variation across cases and whether those diagnostics correspond to prediction reliability.

\subsubsection{Evidence Quality Diagnostics}
\label{sec:sec641}

Table~\ref{tab:flag_trigger_rates_roberta} summarises the evidence-quality flag trigger rates on the 192 holdout cases.

\begin{table}[!htbp]
    \caption{Flag trigger frequencies on the holdout set.}
    \label{tab:flag_trigger_rates_roberta}
    \centering
    \renewcommand{\arraystretch}{1}
    \setlength{\tabcolsep}{8pt}
    \begin{tabular*}{\textwidth}{@{\extracolsep{\fill}}>{\raggedright\arraybackslash}p{0.58\textwidth}>{\raggedleft\arraybackslash}p{0.16\textwidth}>{\raggedleft\arraybackslash}p{0.14\textwidth}@{}}
        \toprule
        \textit{Flag} & \textit{Triggered} & \textit{Rate} \\
        \midrule
        \textbf{Thin evidence ($< 5$ articles)} & 0 & 0.0\% \\
        \textbf{Weight concentration ($\mathrm{HHI} > 0.4$)} & 0 & 0.0\% \\
        \textbf{Label-margin ($< 0.15$)} & 22 & 11.5\% \\
        \textbf{Label-margin skipped (N/A)} & 125 & 65.1\% \\
        \textbf{Source diversity ($> 60\%$ one source)} & 187 & 97.4\% \\
        \textbf{Flip sensitivity ($\leq 5$ articles)} & 68 & 35.4\% \\
        \textbf{Horizon coverage (lookback gap)} & 40 & 20.8\% \\
        \bottomrule
    \end{tabular*}
\end{table}

The most informative diagnostics are flip sensitivity, label margin, and horizon coverage. Thin evidence and weight concentration never trigger, showing that most predictions are not based on too few articles or a single dominant article. Source diversity triggers in almost all cases due to limitations in Finnhub's free-tier outlet coverage, so it is most useful when interpreted alongside other flags.

\begin{figure}[!htbp]
    \centering
    \includegraphics[width=0.70\linewidth]{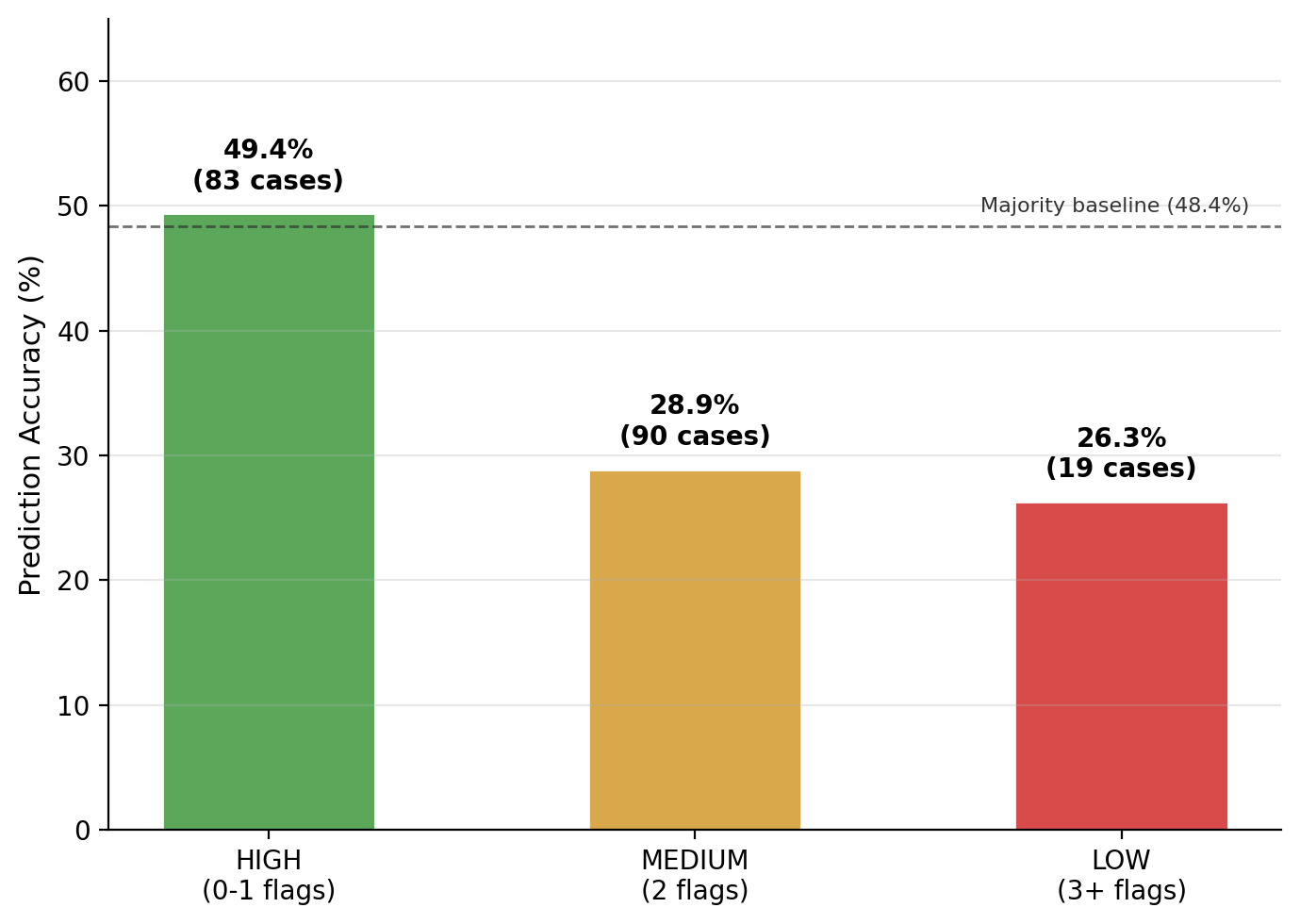}
    \caption{Prediction accuracy by evidence quality rating on the holdout set.}
    \label{fig:quality_acc}
\end{figure}

Figure~\ref{fig:quality_acc} shows that evidence quality ratings correspond to clear accuracy differences. HIGH-rated cases achieve 49.4\% accuracy, slightly exceeding the majority-class baseline, while MEDIUM and LOW cases fall to 28.9\% and 26.3\%, respectively. This suggests that the diagnostic layer provides practical value by identifying when the prediction is more or less reliable.

\subsubsection{Flip-Set Sensitivity}

The flip-set analysis estimates how many articles must be removed before the prediction changes. Table~\ref{tab:flipset_statistics_roberta} summarises the results.

\begin{table}[!htbp]
\centering
\caption{Flip-set statistics on the holdout set.}
\label{tab:flipset_statistics_roberta}
\renewcommand{\arraystretch}{1}
\setlength{\tabcolsep}{8pt}
\begin{tabularx}{\textwidth}{
>{\raggedright\arraybackslash\bfseries}X
>{\raggedleft\arraybackslash}p{3.0cm}}
\toprule
\textit{Statistic} & \textit{Value} \\
\midrule
Median flip-set size & 10 \\
Mean flip-set size & 18.3 \\
Single-article flips & 20 (10.4\%) \\
Unflippable cases & 0 (0.0\%) \\
Cases requiring $> 10$ removals & 92 (47.9\%) \\
\bottomrule
\end{tabularx}
\end{table}

The median prediction requires removing 10 articles before the label changes, and nearly half of cases require more than 10 removals. This indicates that the weighted aggregation generally distributes influence across multiple articles rather than relying on a single dominant source. However, 10.4\% of cases flip after removing one article, identifying fragile predictions that should be treated cautiously.

\subsubsection{Narrative Consistency}
\label{sec:sec643}

To assess the stability of generated explanations, the narrative synthesis component is run five times on 30 holdout cases spanning all prediction windows and five sectors. Across 150 runs, the hallucination validation step detects zero violations. Pairwise cosine similarity between outputs for the same case averages 0.963, as shown in Table~\ref{tab:sector_similarity_stats}.

\begin{table}[!htbp]
\centering
\caption{Similarity statistics by sector on the holdout set.}
\label{tab:sector_similarity_stats}
\renewcommand{\arraystretch}{1.1}
\setlength{\tabcolsep}{8pt}
\begin{tabularx}{\textwidth}{
>{\raggedright\arraybackslash\bfseries}X
>{\raggedleft\arraybackslash}p{1.8cm}
>{\raggedleft\arraybackslash}p{2.5cm}
>{\raggedleft\arraybackslash}p{2.5cm}}
\toprule
\textit{Sector (Company)} & \textit{Cases} & \textit{Avg Similarity} & \textit{Min Similarity} \\
\midrule
Tech (MSFT) & 6 & 0.992 & 0.959 \\
Finance (BAC) & 6 & 0.930 & 0.755 \\
Healthcare (UNH) & 6 & 0.956 & 0.829 \\
Energy (XOM) & 6 & 0.985 & 0.931 \\
Consumer (KO) & 6 & 0.952 & 0.757 \\
Overall & 30 & 0.963 & 0.755 \\
\bottomrule
\end{tabularx}
\end{table}

These results indicate that the low-temperature, evidence-grounded generation setup produces stable summaries. Variation is mainly stylistic, while the core evidence and sentiment attribution remain consistent across repeated runs.

\subsection{Horizon and Sector Analysis}
\label{sec:analysis}

\subsubsection{Performance Across Prediction Horizons}

\begin{figure}[!htbp]
    \centering
    \includegraphics[width=0.95\linewidth]{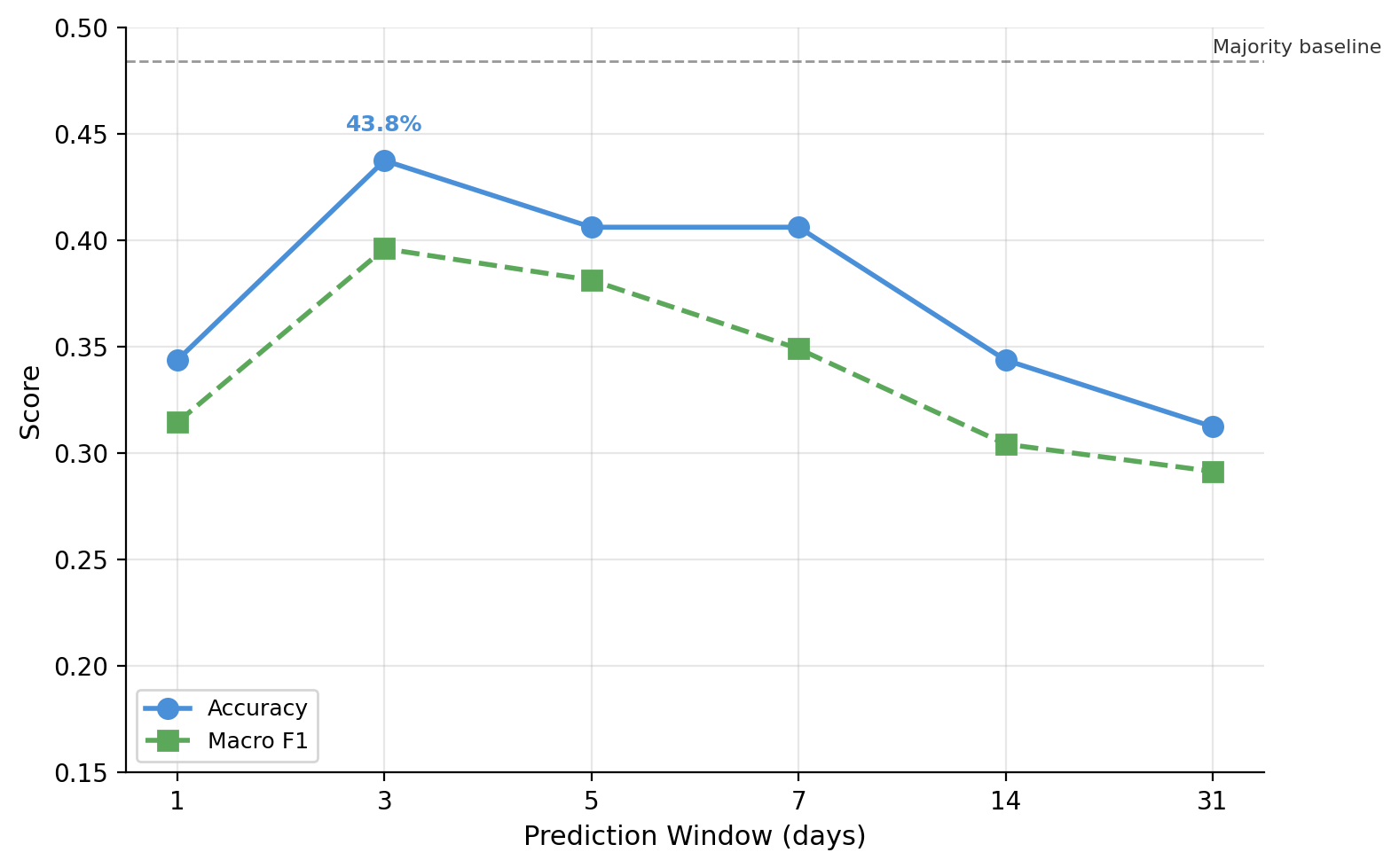}
    \caption{RoBERTa performance across prediction horizons.}
    \label{fig:per_hor}
\end{figure}

Table~\ref{tab:window_comparison_roberta} and Figure~\ref{fig:per_hor} show RoBERTa performance across prediction windows.

\begin{table}[!htbp]
    \caption{Performance across prediction horizons.}
    \label{tab:window_comparison_roberta}
    \centering
    \renewcommand{\arraystretch}{1}
    \setlength{\tabcolsep}{8pt}
    \begin{tabular*}{\textwidth}{@{\extracolsep{\fill}}lcc@{}}
        \toprule
        \textit{Window} & \textit{Accuracy} & \textit{Macro-F1} \\
        \midrule
        \textbf{1 day} & 0.3438 & 0.3146 \\
        \textbf{3 days} & \textbf{0.4375} & \textbf{0.3961} \\
        \textbf{5 days} & 0.4062 & 0.3813 \\
        \textbf{7 days} & 0.4062 & 0.3491 \\
        \textbf{14 days} & 0.3438 & 0.3043 \\
        \textbf{31 days} & 0.3125 & 0.2915 \\
        \bottomrule
    \end{tabular*}
\end{table}

Performance peaks at the 3-day window, followed by the 5-day and 7-day windows. The 1-day window appears too short for stable news aggregation, while 14-day and 31-day windows likely accumulate noise from later events not present in the original article set. The pattern suggests that news sentiment may be most informative over short but not immediate horizons. However, each horizon contains only 32 holdout cases, so these results should be interpreted as indicative rather than statistically conclusive.

\subsubsection{Performance Across Sectors}

Table~\ref{tab:sector_results} reports sector-level holdout performance.

\begin{table}[!htbp]
    \caption{Performance by sector on the holdout set.}
    \label{tab:sector_results}
    \centering
    \renewcommand{\arraystretch}{1}
    \setlength{\tabcolsep}{8pt}
    \begin{tabular*}{\textwidth}{@{\extracolsep{\fill}}>{\raggedright\arraybackslash}p{0.25\textwidth}>{\raggedleft\arraybackslash}p{0.12\textwidth}>{\raggedleft\arraybackslash}p{0.13\textwidth}>{\raggedleft\arraybackslash}p{0.09\textwidth}>{\raggedright\arraybackslash}p{0.25\textwidth}@{}}
        \toprule
        \textit{Sector} & \textit{Accuracy} & \textit{Macro-F1} & \textit{Cases} & \textit{Actual Distribution} \\
        \midrule
        \textbf{Consumer (KO)} & \textbf{0.6250} & 0.2564 & 24 & 1 pos, 8 neg, 15 neu \\
        \textbf{Energy (XOM)} & 0.3333 & 0.2807 & 24 & 14 pos, 4 neg, 6 neu \\
        \textbf{Finance (BAC)} & 0.3333 & 0.2808 & 24 & 7 pos, 3 neg, 14 neu \\
        \textbf{Healthcare (UNH)} & 0.2083 & 0.1629 & 24 & 5 pos, 4 neg, 15 neu \\
        \textbf{Tech (4 companies)} & 0.3750 & \textbf{0.3687} & 96 & 34 pos, 19 neg, 43 neu \\
        \bottomrule
    \end{tabular*}
\end{table}

Performance varies substantially across sectors. Consumer obtains the highest accuracy, largely because the model frequently predicts neutral for Coca-Cola. Technology obtains the highest macro-F1 across 96 cases, suggesting more balanced performance. Healthcare performs worst, likely because news sentiment in this sector is mediated by regulatory, legal, and policy factors whose market impact is difficult to infer from linguistic tone alone. Since most sectors are represented by one holdout company, these results should be interpreted cautiously.

\subsection{Lessons from Zero-Shot Financial Prediction}

The results suggest three main lessons. First, strong sentiment classification does not imply strong stock prediction. FinGPT and FinBERT perform very well on Financial PhraseBank, but this advantage does not carry over to the pipeline-level prediction task. This supports the distinction between recognising sentiment in text and forecasting subsequent market movement.

\begin{figure}[!htbp]
    \centering
    \includegraphics[width=0.60\linewidth]{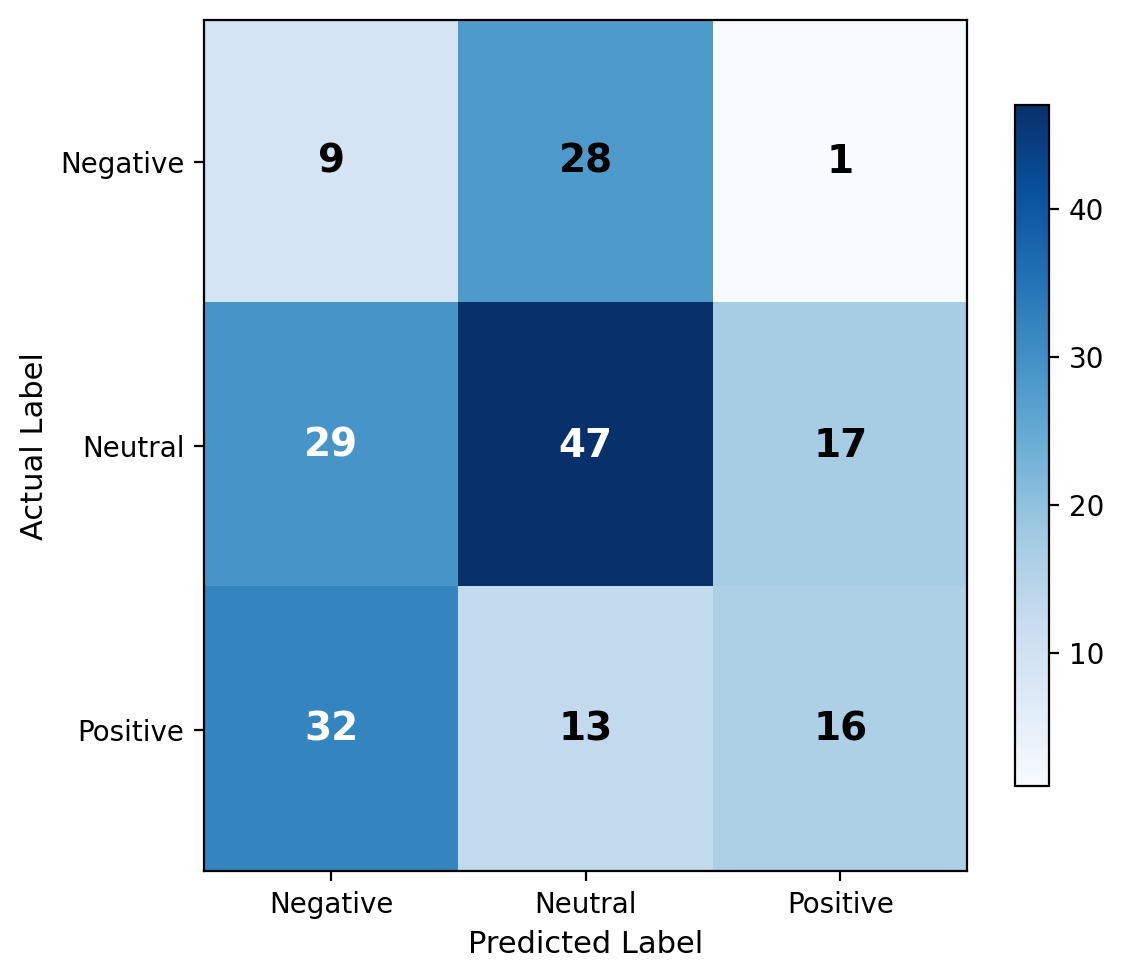}
    \caption{RoBERTa confusion matrix on the holdout set.}
    \label{fig:conf}
\end{figure}

Second, negative stock movements are particularly difficult to predict from news sentiment. The argmax models each correctly identify only one of 38 negative cases, while RoBERTa with decision thresholds identifies 9 of 38. Figure~\ref{fig:conf} shows the RoBERTa confusion matrix. Although thresholding improves negative recall, most negative cases are still absorbed into neutral.

This failure is not caused by an inability to recognise negative financial language, since all models perform much better on the negative class in Financial PhraseBank. Instead, it reflects the weaker relationship between article-level sentiment and subsequent price direction. Negative stock returns may be driven by broader market conditions, missed expectations, price corrections, or information not present in the retrieved news. Financial news also often expresses negative developments indirectly, using hedged language rather than explicit negative sentiment \citep{GurunButler2012}.

Third, explainability provides value even when accuracy is limited. Evidence quality ratings separate more reliable from less reliable predictions, flip-set analysis exposes fragile cases, and narrative summaries remain stable and grounded. Overall, zero-shot financial NLP does not reliably outperform the majority-class baseline in raw accuracy, but the proposed NLI-based pipeline achieves the best macro-F1 among the tested models and avoids invalid-output failures from generative baselines. These findings suggest that the main value of zero-shot financial NLP in this setting may not be autonomous prediction, but transparent evidence organisation and reliability-aware decision support.

\section{Limitations}
\label{sec:limitations}

This study has several limitations. First, the evaluation depends on the Finnhub free-tier API as the sole news source. Although Finnhub provides structured access to company news, the free tier offers limited historical coverage, restricted outlet diversity, and variable article availability. This affects both the breadth and consistency of the evidence available to the pipeline. In particular, repeated or syndicated articles can dominate the retrieved article set, while relevant company-specific news may be missing from some lookback windows. As a result, some of the observed prediction difficulty may reflect limitations in input coverage rather than only limitations of the models.

Second, the evaluation is restricted to English-language news for US-listed companies. The results may not generalise to international markets, non-English news sources, or companies with lower media visibility. These settings may differ substantially in news volume, reporting style, publication timing, and market response dynamics.

Third, the pipeline uses only textual evidence from news articles. It does not incorporate complementary market variables such as price momentum, trading volume, volatility, analyst revisions, sector indices, or macroeconomic indicators. The substantial overlap between score distributions across ground-truth classes suggests that news sentiment alone may be insufficient for reliable short-horizon directional prediction. This is especially relevant for negative movements, which may be driven by broader market conditions, missed expectations, or information channels not captured in the retrieved news.

Fourth, the evaluation covers a nine-month period and 20 companies. Although the dataset spans multiple sectors and prediction horizons, financial markets are non-stationary, and news-price relationships may change across volatility regimes, earnings cycles, and macroeconomic conditions. Sector-level conclusions should also be interpreted cautiously, since most sectors in the holdout set are represented by a single company.

Finally, the ground-truth labelling depends on a volatility-scaled neutral threshold with a fixed scaling constant of $k=0.5$. This threshold provides a practical way to distinguish meaningful movement from noise, but alternative values would produce different class distributions. Since the threshold defines the prediction task itself, there is no single objective boundary between neutral and directional returns.

\section{Conclusion}
\label{sec:conclusion}

This paper investigated whether financial news can support short-horizon stock movement prediction using zero-shot financial NLP. We introduced an impact-aware pipeline that combines company relevance filtering, recency weighting, event-dependent impact-horizon weighting, zero-shot NLI classification, and multi-layered explainability, and evaluated it on 480 cases across 20 US-listed companies, six prediction horizons, and five sentiment models. The results show that zero-shot financial NLP remains limited for direct stock movement prediction: none of the tested models exceeded the majority-class baseline on holdout accuracy, and all struggled particularly with negative movements. This suggests that the main difficulty is not simply sentiment classification, since the models perform substantially better on clean Financial PhraseBank examples, but the weaker and noisier relationship between article-level sentiment and subsequent price direction. At the same time, the proposed pipeline provides value beyond raw accuracy: the NLI-based configuration achieved the strongest macro-F1 among the tested models, avoided invalid-output failures observed in generative LLM baselines, and produced reliability signals through evidence quality ratings, flip-set analysis, and grounded narrative summaries. Overall, the findings suggest that financial news should not be treated as a standalone source of reliable directional trading signals, but as structured evidence for decision support. By exposing which articles contribute to a prediction, how temporally aligned they are, how robust the aggregated decision is, and whether the evidence base is trustworthy, the proposed framework provides a transparent way to understand when zero-shot financial NLP succeeds, when it fails, and why.

\section{Competing Interests}

On behalf of all authors, the corresponding author states that there is no conflict of interest.

\section{Funding Information}

This research received no specific grant from any funding agency in the public, commercial, or not-for-profit sectors.

\section{Author Contribution}

Hanrui Luo was responsible for writing the code and experiments along with drafting the initial version of the paper. Shreyank N Gowda, was responsible for supervision of the project, along with guiding Hanrui on the direction of the project.

\section{Data Availability Statement}
All data used in this paper is from publicly available resources and these have been listed in the paper. We will be releasing the code on acceptance and this would include steps to download and reproduce our results.

\section{Research Involving Human and/or Animals}
Not applicable.

\section{Informed Consent}
Not applicable.

\nocite{*}
\bibliography{main}

\end{document}